\documentclass[11pt,a4paper]{article}
\usepackage[hyperref]{acl2021}
\usepackage{times}
\usepackage{latexsym}
\usepackage{graphicx}
\usepackage{subfigure}
\usepackage{url}
\usepackage{booktabs}
\usepackage{multirow}
\usepackage{amsmath}
\usepackage{amsfonts}
\usepackage{diagbox}

\usepackage{color}

\usepackage{microtype}

\aclfinalcopy % Uncomment this line for the final submission
\title{\emph{Addressing Inquiries about History}: An Efficient and Practical Framework for Evaluating Open-domain Chatbot Consistency}

\author{
    Zekang Li$^1$$^2$, Jinchao Zhang$^3$, Zhengcong Fei$^1$$^2$, Yang Feng$^1$$^2$\thanks{\ \ Joint work with Pattern Recognition Center, WeChat AI, Tencent Inc. Yang Feng is the corresponding author. Work was done when Zekang Li and Zhengcong Fei were intern at WeChat AI.}, \bf{Jie Zhou}$^3$ \\
  $^{1}$ Key Laboratory of Intelligent Information Processing \\
  Institute of Computing Technology, Chinese Academy of Sciences (ICT/CAS) \\
  $^{2}$ University of Chinese Academy of Sciences \\
  $^{3}$ Pattern Recognition Center, WeChat AI, Tencent Inc, China \\
  {\tt \{\href{mailto:lizekang19g@ict.ac.cn}{lizekang19g},\href{mailto:feizhengcong@ict.ac.cn}{feizhengcong},\href{mailto:fengyang@ict.ac.cn}{fengyang}\}@ict.ac.cn} \\
  {\tt \{\href{mailto:dayerzhang@tencent.com}{dayerzhang},\href{mailto:withtomzhou@tencent.com}{withtomzhou}\}@tencent.com}
  \\}

\date{}

\begin{document}
\maketitle
\begin{abstract}
A good open-domain chatbot should avoid presenting contradictory responses about facts or opinions in a conversational session, known as its consistency capacity. 
However, evaluating the consistency capacity of a chatbot is still challenging.
Employing human judges to interact with chatbots on purpose to check their capacities is costly and low-efficient, and difficult to get rid of subjective bias.
In this paper, we propose the \textbf{A}ddressing \textbf{I}nquiries about \textbf{H}istory (\textbf{AIH}), an efficient and practical framework for the consistency evaluation. 
At the conversation stage, \textbf{AIH} attempts to address appropriate inquiries about the dialogue history to induce the chatbot to redeclare the historical facts or opinions.
We carry out the conversation between chatbots, which is more efficient than the human-bot interaction and can also alleviate the subjective bias.
In this way, we manage to rapidly obtain a dialog session that contains responses with high contradiction possibilities.
At the contradiction recognition stage, we can either employ human judges or a natural language inference (NLI) model to recognize whether the answers to the inquiries are contradictory with history.
Finally, we are able to rank chatbots according to the contradiction statistics. 
Experiments on open-domain chatbots show that our approach can efficiently and reliably assess the consistency capacity of chatbots and achieve a high ranking correlation with the human evaluation.
We release the framework and hope to help improve the consistency capacity of chatbots.\footnote{\url{https://github.com/ictnlp/AIH}}
\end{abstract}

\section{Introduction}
\begin{table}[t]
		\centering
		\small
		\begin{tabular}{rp{5.5cm}}
			\toprule[1pt]
			\multicolumn{2}{c}{\bf DialoGPT 762M} \\
			\hline
			\bf{Human:} & \textit{What movies do you like most?} \\
			\bf{DialoGPT:} & \textit{The \textcolor{red}{Unforgiven} from Metallica.} \\
			\bf{Human:} & \textit{What do you think of the Unforgiven?} \\
			\bf{DialoGPT:} & \textit{\textcolor{red}{I never heard of it.}} \\
			\bottomrule[1pt]
			\multicolumn{2}{c}{\bf Blender 2.7B} \\
			\hline
			\bf{Human:} & \textit{What do you like to cook?} \\
			\bf{Blender:} & \textit{\textcolor{red}{I only cook Indian cuisine.} How about you?} \\
			\bf{Human:} & \textit{I enjoy cooking Chinese food, especially the dumplings.} \\
			\bf{Blender:} & \textit{Chinese food is delicious. \textcolor{red}{I also like cooking the Chinese food.}} \\
			\bottomrule[1pt]
			\multicolumn{2}{c}{\bf Plato 1.6B} \\
			\hline
            \bf{Human:} & \textit{Do you like dogs?} \\
			\bf{Plato:} & \textit{Yes, dogs are cool, I prefer cats though.} \\
			\bf{Human:} & \textit{Do you have a dog?} \\
			\bf{Plato:} & \textit{\textcolor{red}{I do not, but I'd love a dog.} I think dogs are so cute.} \\
			\bf{Human:} & \textit{I love dogs very much. I have three dogs.} \\
			\bf{Plato:} & \textit{\textcolor{red}{I have three dogs too.} One puppy and two Huskies.} \\
			\bottomrule[1pt]
		\end{tabular}
		\caption{Several human-bot conversations demonstrate that popular chatbots (DialoGPT, Blender, and Plato) generate inconsistent responses when talking to a human under some specific conditions. }
		\label{tab:no-consist-example}
		%\ {-1\baselineskip}
	\end{table} 
	
In recent years, open-domain intelligent chatbots \cite{adiwardana2020towards,bao2020plato,smith2020can} have approached great progress due to the development of the large-scale pre-training approaches  \cite{devlin2019bert,radford2019language,brown2020language} and the large amount of high-quality conversational datasets \cite{dinan2018wizard,baumgartner2020pushshift,smith2020can}. 
Though the success is indisputable and exciting, there is still a long way to build a truly human-like open-domain chatbot. 

Current open-domain chatbots hold a superiority in generating fluent, engaging, and informative responses, but show the soft spot on consistency \cite{nie2020like}. As shown in Table \ref{tab:no-consist-example}, we present some interactive dialogue samples between human and several popular open-domain chatbots (e.g. DialoGPT \cite{zhang2019dialogpt}, Blender \cite{smith2020can}, and Plato \cite{bao2020plato}). 
All open-domain chatbots occasionally generate responses that are contradictory with history when interacting with humans, which is really annoying and severely disrupts the communication once happening. Therefore, it is imperative to improve the consistency of the open-domain chatbots. However, one crucial reason that restricts consistency development is the lack of an effective and practical evaluation method. 

% human bot 交互time and cost inefficient
To estimate the consistency of chatbots, the most straightforward approach is to ask human annotators to distinguish whether the conversations generated from the chatbots are consistent or not. 
However, the instructions followed by annotators are often chosen ad-hoc, and there is no explicit definition, which leads to the relatively low inter-agreement in the human chatbot consistency evaluation \cite{mehri2020unsupervised}. As a result, several works have been proposed to develop automatic evaluation methods \cite{welleck2020dialogue,song2020profile,nie2020like}. %\citet{welleck2020dialogue,song2020profile} focused on the persona-related consistency and profile-related consistency and characterized the chatbot consistency evaluation as natural language inference problems. \citet{nie2020like} proposed a new human-craft dataset and the structured utterance-based approach to detect the contradictions with dialogue history. 
While these methods can detect contradictions efficiently in the dialogue, they depend on the human-bot conversations, which is still cost-inefficient and tend to suffer from low quality \cite{deriu2020spot,dinan2020second}. Besides, the occurrence rate of contradiction is low under this condition. All these problems slow down the development of  consistency evaluation of dialogue systems severely. 

% 在human bot交互中，一致性问题比较随机，出现的概率较少，低效，并且标注人员的一致性较低。

Towards that end, based on the observations: (i) chatbots are likely to generate contradictions when chatting about facts and opinions; (ii) answering the  questions about the conversational history correctly can reveal the ability to understand the conversation and keep consistency, we present the \textit{Addressing Inquiries about History} (AIH) framework, an effective and practical framework for open-domain chatbot consistency evaluation. 
The framework can be used to rank different chatbots with regard to the ability to be consistent with themselves in the conversation. 
Specifically, AIH consists of two stages: (i) during the inquiry stage,  questions about the facts and opinions mentioned in the conversation history are inserted into the conversation between chatbots; (ii) during the contradiction recognition stage,  the responses of the inserted questions are collected, and automatic models or human judges can be adopted to decide whether the responses are consistent with the dialogue history.  

In brief, our AIH has the following key advantages: \textit{Firstly}, it is based on bot-bot conversation, which avoids the human intervention and brings down the cost and time effort significantly.  \textit{Secondly}, by inserting specific questions, contradictions occur more frequently, and it is easier for human annotators or automatic consistency detection model to distinguish the contradiction compared with natural conversations.
Extensive experiments demonstrate that the proposed framework can produce effective, efficient, and reliable consistency evaluation. Furthermore, we also make an in-depth discussion about the influence of question generation, contradiction detection, and evaluation agreement in our framework. 
%We apply the framework to several popular open-domain chatbots and produce an accurate and stable consistency ranking among them. Furthermore, we make more in-depth discussion about the effectiveness and limitation of different modules in our framework. We hope \textit{Addressing Inquiries} can facilitate and provide standard evaluation for the future work on developing self-consistent open-domain chatbots.

Our contributions are summarized as follows:
\begin{itemize}
    \item We propose the \textit{Addressing Inquiries about History}(AIH), an effective and practical framework for open-domain chatbot consistency evaluation.
    \item Experiments show that AIH can produce effective, efficient, and reliable consistency evaluation.
    We release the framework as a ready-to-use tool for evaluating the consistency of chatbots. We hope AIH can facilitate and provide standard evaluation for future work on developing self-consistent open-domain chatbots.
\end{itemize}

\begin{figure*}[t]
    \centering
    \includegraphics[width=0.95 \textwidth]{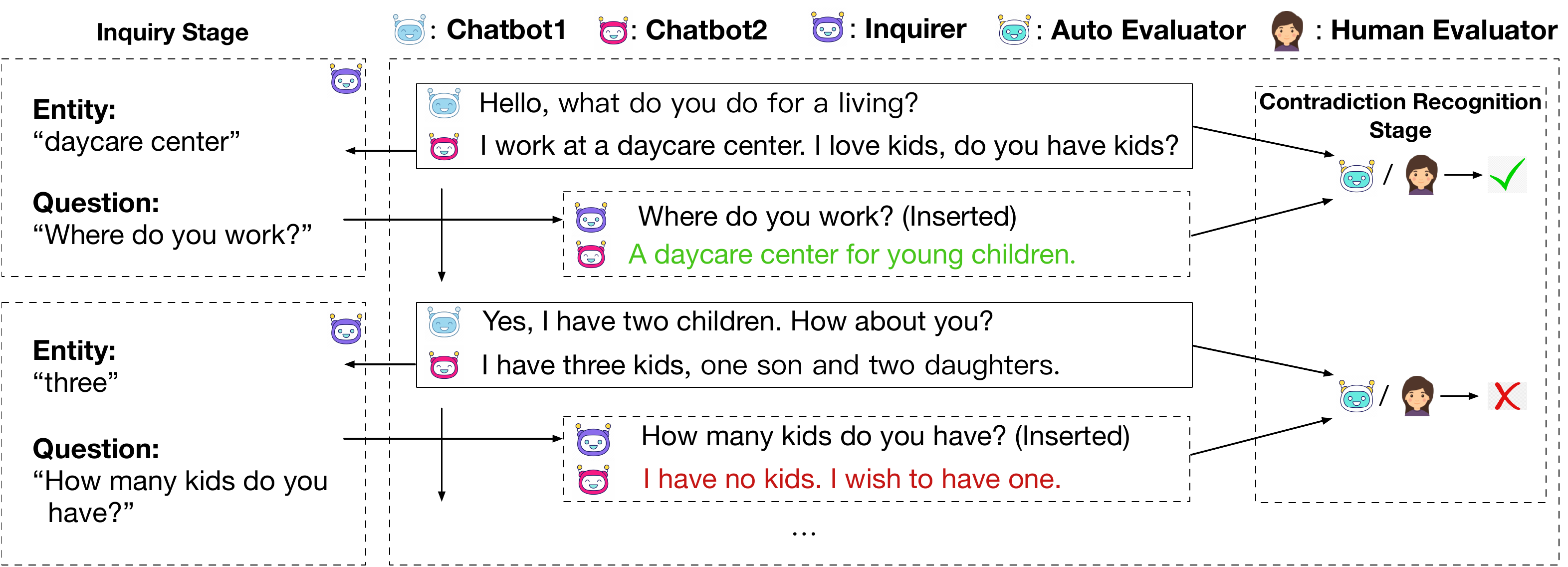}
    \caption{Overview of the Addressing Inquiries about History framework. There are five “agents”: \textbf{Chatbot1}, \textbf{Chatbot2}, \textbf{Inquirer}, \textbf{Auto Evaluator}, and \textbf{Human Evaluator} and two stages: \textbf{Inquiry Stage} and \textbf{Contradiction Recognition Stage} in the framework. Chatbot1 and Chatbot2 are the participants in the bot-bot conversation, in which Chatbot2 is the one to be evaluated. In the Inquiry stage, Inquirer extracts opinion- or fact-related entities and generate inquiries based on them. In the Contradiction Recognition stage, the Auto Evaluator is generally a contradiction detection model to automatically evaluate if the responses from Chatbot2 are consistent and the Human Evaluator can also be employed for more accurate evaluation. Note that the inserted inquiries do not affect the natural conversation. Better view in color. }
    \label{fig:framework}
\end{figure*}

\section{Related Work}
% \subsection{Open-domain Chatbot Evaluation}
There are various methods to evaluate the consistency of chatbots, containing automatic and human-based methods. The methods mainly fall into two dimensions: the static and interactive evaluation. 

\subsection{Static Evaluation} Static evaluation denotes evaluating if the responses generated based on the static context are contradictory with the pre-defined persona or profile and the dialogue history by neural models or human annotators. 
\citet{welleck2020dialogue} and \citet{song2020profile} focus on the persona-related consistency and profile-related consistency, and characterize the chatbot consistency evaluation as the natural language inference problem. 
\citet{nie2020like} build a new human-craft dataset called DECODE and propose a structured utterance-based approach to detect the contradictions in the dialogue history. While being cost-efficient, static evaluation can not accurately reflect the conversation capacity of the chatbot in the real world.

\subsection{Interactive Evaluation}
\noindent \textbf{Human-bot Conversations.} In order to pursue more authentic evaluation, the standard method is to let humans converse with a chatbot and evaluate it by aforementioned models or humans afterward \cite{mehri2020unsupervised}. However, apart from the high cost of collecting human-bot conversations, there is also a high cognitive strain on humans, which leads to unstable results \cite{dinan2020second}. 

\noindent \textbf{Bot-bot Conversations.} Recently, bot-bot conversations, which significantly reduce the cost and human bias, are focused. \citet{deriu2020spot,DBLP:journals/corr/abs-1909-03087} propose to let humans evaluate bot-bot or self-talk conversations to give a relative ranking of the overall quality of chatbots. 
Different from these methods, we focus on the chatbot consistency and insert inquiries to redeclare historical facts. And we introduce both automatic and human approaches to evaluate the chatbot consistency. 

% \subsection{Conversational Contradictions Detection}
% Several prior works on detecting contradictions in dialogue 

\section{Approach}
In this section, we first provide an overview of the \emph{Addressing Inquiries about History} (AIH) framework. We then describe the Inquiry stage, the Contradiction Recognition stage, and the chatbot ranking process.

\subsection{Overview}

To estimate the consistency capacity, questions about the opinions and facts in the dialogue history are inserted into the current bot-bot conversation. Then, the corresponding responses are collected and judged by automatic tools or human evaluation. The workflow of our proposed  AIH framework is shown in Figure \ref{fig:framework}. 

%As shown in Figure \ref{fig:framework}, Addressing Inquiries framework works by inserting questions about the personality and facts in the dialogue history into the bot-bot conversation, then collecting the responses and using neural models or letting human judges to decide for each response if it is consistent with the dialogue history. 

To be specific, there are five “agents” in the framework:  Chatbot1, Chatbot2,  Inquirer, Auto Evaluator, and Human Evaluator. The Chatbot1 and Chatbot2 are the participants in the bot-bot conversation. The Inquirer extracts opinion- or fact-related entities and generates inquiries based on the entities. The Auto Evaluator is generally a contradiction detection model to automatically evaluate if the responses from Chatbot2 are consistent. The Human Evaluator is used for more accurate evaluation.

Formally, assume a pool of $N$ chatbots $\{B_1, ..., B_N\}$ which are ready to be evaluated in terms of consistency capacity. For each pair of chatbots (referred as Chatbot1 and Chatbot2), we let Chatbot1 talk with Chatbot2 for $K$ turns. Note that Chatbot2 is the one to be evaluated.
($i$) During the inquiry stage, within the conversation between Chatbot1 and Chatbot2, for each utterance $u_{2k}$ generated by Chatbot2, Inquirer extracts the entities about opinions and facts, then asks Chatbot2 a question $q_k$ about these entities, where $k$ is the turn number. Chatbot2 answers the question $q_k$ and generates the corresponding response $r_k$. Note that we ignore the questions generation operation when there is no entity that can be extracted.
($ii$) During the contradiction recognition stage, we use neural models (e.g. Natural Language Inference Model) or employ human judges to decide if the utterance pair $\{u_{2k}, r_k\}$ exists non-consistent problem.
We collect at least $M$ dialogues from each chatbot pair, then calculate the ranking order on the consistency. In this way, we can discriminate the consistency capability of chatbots effectively and efficiently. 
In the following, we will introduce the inquiry stage and the contradiction recognition stage, respectively. 

\subsection{Inquiry Stage}
Based on our observation and prior work \cite{nie2020like}, in natural human-bot or bot-bot conversation, contradiction is more likely to occur when chatting about repeated facts and opinions, especially after similar questions. 
Therefore, to mimic such a contradiction occurrence process, we make chatbots to produce responses by asking chatbots related questions about previous facts and opinions. 
In this condition, generating  appropriate questions is pretty important. Hence, we first extract entities about facts and opinions from the historical utterances, then employ a neural model to generate questions about the extracted entities.

\noindent \textbf{Entity Extraction} \\
Considering that chatbots usually generate contradictions when chatting about facts and opinions,
we apply Named Entity Recognition tools in Stanza \cite{qi2020stanza}, a popular natural language analysis package, to extract named entities from utterance $u_{2k}$ containing person, organization, location, etc. \footnote{There are 18 named entity types. Please refer to \cite{weischedel2013ontonotes} for more details.} For example, for the utterance \textit{“I would love to visit New York next year.”}, we can extract out two entities: \textit{“New York”} and \textit{“next year”}.

\noindent \textbf{Question Generation Model} \\
For question generation, we employ UniLM \cite{dong2019unified} model that is fine-tuned on the SQuAD dataset \cite{rajpurkar2016squad} with question generation task \cite{text2text@2020}. We utilize a public implementation and checkpoint.\footnote{\url{https://github.com/artitw/text2text}}
In our framework, given the entities extracted before and utterance, UniLM generates a suitable question for each entity.  For example, given \textit{“New York”} and \textit{“I would love to visit New York next year.”}, the model generates \textit{“Where would you like to visit next year?”}. We then randomly select one question and insert it into the bot-bot conversation.

\subsection{Contradiction Recognition Stage}
In our framework, since the question $q_k$ is based on the previous Chatbot2's utterance $u_{2k}$, the response from Chatbot2 should be consistent with the utterance $u_{2k}$. Therefore, the Auto Evaluator and Human Evaluator can just consider the answer $r_k$ and utterance $u_{2k}$. \\ 
\noindent \textbf{Auto Evaluator} \\
For automatic evaluation, the Auto Evaluator is generally a contradiction detection model. The Auto Evaluator take the response $r_k$ answered by Chatbot2 and the previous utterance $u_{2k}$ as input, and output the contradiction score $y_{k}$. It can be formulated as:
\begin{equation}
    y_{k} = f_{\theta}(r_k, u_{2k}),
\end{equation}
where $f_\theta$ is the detection function and $\theta$ is the parameters.
Compared to other contradiction detection methods that consider the whole dialogue, the Auto Evaluator can refrain from the noise contained in the whole dialogue. In practice, we select the Roberta-large model \cite{liu2019roberta} fine-tuned on the Multi-Genre Natural Language Inference dataset \cite{N18-1101} as the implementation of Auto Evaluator.\footnote{\url{https://huggingface.co/roberta-large-mnli}}

\noindent \textbf{Human Evaluator} \\
In traditional dialogue consistency evaluation methods, human judges are asked to read the whole dialogue and give an overall consistency score, usually 0 or 1. In our opinion, these methods suffer from high cost and low inter-agreement because there is no specific instruction, and it is too hard for human judges to give an overall score on the whole dialogue \cite{mehri2020unsupervised}. 

In our framework, human evaluators are only asked to decide if the response $r_k$ answered by Chatbot2 is consistent with the previous utterance $u_{2k}$ or not, which is more specific and easier than the traditional methods. As a result , the cost decreases, and the evaluation quality increases.  Besides,  the human annotation in our framework is much more fine-grained than the traditional methods, which can provide more information for the development cycle of dialogue systems.

\subsection{Consistency Metrics and Bot Ranking}
Based on the previous results, we can obtain a ranking list about different chatbots on consistency capacity. 
Formally, for each chatbot pair $\{B_i, B_j\}$, we collect $M$ dialogues. For each inquiry pair, the detection of contradiction is made by comparing $y_{k}$ with a threshold $\tau$:
\begin{equation}
    c_k = \mathbb{I}(f_{\theta}(r_k, u_{2k}) > \tau). \label{tau}
\end{equation}
The contradiction rate of the chatbot $B_j$ within chatbot pair $B_{ij}$ can be computed as:
\begin{equation}
    C_{ij} = \frac{1}{M}\sum^{m} c_k,
\end{equation}
where $m$ is the number of inquiries in each dialog and $M$ is the total number of inquiry pairs.
For the overall contradiction rate of the chatbot $B_j$ is calculated as:
\begin{equation}
    C_j = \frac{1}{N} \sum_{i=1}^{N} C_{ij}.
\end{equation}
Finally, we can rank the chatbots using the overall contradiction rate.

\section{Experiment Setup}
In this section, we first list the dialogue systems used in our experiments, then describe the experimental settings in detail.
\subsection{Chatbots}
We select several popular open-domain chatbots in our experiments. \\
\noindent \textbf{Blender (BL)} \cite{roller2020recipes} is firstly pre-trained on Reddit dataset \cite{baumgartner2020pushshift} and then fine-tuned with high-quality human annotated dialogue datasets (BST), which containing four datasets: Blended Skill Talk \cite{smith2020can}, Wizard of Wikipedia \cite{dinan2018wizard}, ConvAI2 \cite{dinan2020second}, and Empathetic Dialogues \cite{rashkin2018towards}. By fine-tuning, Blender can learn blend conversational skills of engagement, knowledge, empathy and personality. Blender has three model sizes: 90M, 2.7B, and 9.4B. Since 2.7B parameter model achieves the best performance in \cite{roller2020recipes}, we use the 2.7B version in our experiments.\\
\noindent \textbf{Plato (PL)} \cite{bao2020plato} is an open-domain chatbot, pre-trained on Reddit dataset and fine-tuned with BST dataset, which is claimed to be superior to Blender. According to the evaluation in \cite{bao2020plato}, we select the 1.6B parameter version in our experiments.  \\
\noindent \textbf{DialoGPT (DG)} \cite{zhang2019dialogpt} is trained on the basis GPT-2 \cite{radford2019language} using Reddit comments. There are three model sizes: 117M, 345M, and 762M. We fine-tuned the 762M version on the BST datasets.  \\
\noindent \textbf{DialoFlow (DF)} \cite{li2021wechat,li2021dialoflow} is a top method in DSTC9 Interactive Dialogue Evaluation track \cite{gunasekara2020overview}. We reproduced the DialoFlow model based on GPT2-large \cite{radford2019language} and fine-tuned it with BST dataset.  \\

\subsection{Experimental Settings} 

We adopt four experimental paradigms to evaluate the effectiveness of the AIH. \\
\noindent \textbf{Bot-Bot Interaction.} For bot-bot interaction, the maximum interaction turn is set to 15. All chatbots exploit Nucleus Sampling \cite{holtzman2019curious} with $p$=0.9 when generating responses. For each chatbot pair, we collect at least 200 dialogues. 

\noindent \textbf{Human Annotation.} To verify the effectiveness of our framework, we conduct the human evaluation. For the bot-bot conversation under our framework, we employ three professional human annotators from a commercial data annotating company to separately annotate three fields: whether inquiry chatbot generates appropriate questions, whether Chatbot2 answers the questions relevantly, and whether the responses from chatbot2 are contradictory with the dialogue history. We pay the company reasonable salary. The company provides comfortable working conditions and fair salaries for the annotators.
For each chatbot pair, we randomly sample 50 dialogues to be annotated. We compute the final decision via voting.

In Human-bot Natural Interaction and Expert Evaluation, we deployed the four chatbots on the remote server and designed a web interface. Human could chat with a random chatbot each time through the web interface and give the consistency score, being unaware of which chatbot they are chatting with.

\noindent \textbf{Human-Bot Natural Interaction.}  For each chatbot, we collect dialogues by inviting student volunteers from our university through the web interface. The participants were instructed to chat with the chatbots naturally. They were asked to sign the agreement before the conversation: ($i$) They were at least 18 years old and could type into the web interface to chat with the chatbots. ($ii$) They were told that their interaction would generate anonymous text data which would only be used for the research on dialogue systems. 
We filtered out the dialogues with $\textless$5 turns and the dialogues with abusive words. For each chatbot, there are at least 40 eligible dialogues. Then we employ the three professional human annotators to individually annotate whether each utterance from the chatbot is consistent or not.  

\noindent \textbf{Expert Evaluation.} To obtain the human ranking for the consistency of the chatbots, we invite three expert volunteers from our lab, who have 2-3 years experience of dialogue system development, to chat with each bot at least 10 times and about 15 turns each time.  In the chatting, experts are asked to intentionally induce the chatbots to re-answer the questions about the dialogue history and give the consistency score from 0 to 1. Note that we ask the experts to chat with the chatbots for $\textgreater$20 times before the formal evaluation. We average the scores from three experts as the overall consistency score.

Note that Expert Evaluation and Human Annotation were done before the automatic evaluation. Human-bot Natural Interaction was done after the automatic evaluation. All human evaluations were independent from the automatic evaluation.

\section{Experimental Results} 
In this section, we conduct experiments to illustrate the effectiveness, efficiency, and stability of the proposed AIH framework.

\begin{table}[t]
		\centering
		\begin{tabular}{c|cccc}
			\toprule[1pt]
			\multicolumn{5}{c}{Expert Consistency Score $\uparrow$} \\
			\hline
			 & \bf BL & \bf PL & \bf DG & \bf DF \\
			\hline
			\bf{Expert.1} & 0.55 & 0.80 & 0.72 & 0.69 \\
			\bf{Expert.2} & 0.37 & 0.87 & 0.60 & 0.56 \\
			\bf{Expert.3} & 0.31 & 0.89 & 0.60 & 0.55 \\
			\hline
			\bf{Avg.} & 0.41 & \bf 0.85 & 0.64 & 0.60 \\
			\bottomrule[1pt]
		\end{tabular}
		\caption{The expert consistency score of each chatbot. Higher is better.}
		\label{tab:expert_score}
		%\ {-1\baselineskip}
\end{table}

\begin{table}[t]
		\centering
		\begin{tabular}{c|cccc|c}
			\toprule[1pt]
% 			\multicolumn{6}{c}{Contradiction Rate (Auto $\tau=0.5$) $\downarrow$} \\
% 			\hline
% 			 & \bf BL & \bf PL & \bf DO & \bf DF & \bf Avg. \\
% 			\hline
% 			\bf{BL} & 0.295 & 0.166 & 0.257 & 0.265 & \bf 0.246 \\
% 			\bf{PL} & 0.282 & 0.178 & 0.224 & 0.241 & 0.231 \\
% 			\bf{DO} & 0.294 & 0.166 & 0.241 & 0.230 & 0.233 \\
% 			\bf{DF} & 0.284 & 0.169 & 0.237 & 0.250 & 0.235 \\
% 			\hline
% 			\bf{Avg.} & 0.289 & \bf 0.170 & 0.239 & 0.246 & 0.236 \\
			\multicolumn{6}{c}{Contradiction Rate (Auto $\tau=0.15$) $\downarrow$} \\
			\hline
			 & \bf BL & \bf PL & \bf DG & \bf DF & \bf Avg. \\
			\hline
			\bf{BL} & 0.431 & 0.240 & 0.324 & 0.362 & 0.339 \\
			\bf{PL} & 0.431 & 0.263 & 0.293 & 0.357 & 0.336 \\
			\bf{DG} & 0.425 & 0.251 & 0.344 & 0.345 & 0.341 \\
			\bf{DF} & 0.427 & 0.264 & 0.344 & 0.371 & 0.351 \\
			\hline
			\bf{Avg.} & 0.428 & \bf 0.255 & 0.326 & 0.359 & 0.342 \\
			\toprule[1pt]
			\multicolumn{6}{c}{Contradiction Rate (Human) $\downarrow$} \\
			\hline
			 & \bf BL & \bf PL & \bf DG & \bf DF & \bf Avg. \\
			\hline
			\bf{BL} & 0.487 & 0.282 & 0.398 & 0.396 & 0.391 \\
			\bf{PL} & 0.411 & 0.212 & 0.500 & 0.435 & 0.390 \\
			\bf{DG} & 0.404 & 0.211 & 0.304 & 0.431 & 0.338 \\
			\bf{DF} & 0.462 & 0.268 & 0.310 & 0.377  & 0.354 \\
			\hline
			\bf{Avg.} & 0.441 & \bf 0.243 & 0.378 & 0.410 & 0.368 \\
			\bottomrule[1pt]
		\end{tabular}
		\caption{The contradiction rate of each chatbot pair. The column name and the row name represent Chatbot1 and Chatbot2 respectively.}
		\label{tab:avg_contradiction_rate}
		%\ {-1\baselineskip}
\end{table}

\subsection{Evaluation Effectiveness}
We report the expert ranking results in expert evaluation, automatic evaluation, and human evaluation under the AIH framework, respectively. %which demonstrates that our framework can effectively evaluate the chatbot consistency and the evaluation results agree with human experts on ranking. \\

\noindent \textbf{Expert Ranking.}
Table \ref{tab:expert_score} shows the expert consistency scores for different chatbots. We can find that Plato achieves the best expert consistency score, up to 0.85. And the ranking of consistency for these four chatbots is: Plato $\textgreater$ DialoGPT $\textgreater$ DialoFlow $\textgreater$ Blender, which can serve as the gold reference.

\noindent \textbf{Auto Evaluation Results.}
Table \ref{tab:avg_contradiction_rate} shows the contradiction rate of each chatbot pair in auto evaluation. The lower contradiction rate means the better consistency. The column name and the row name represent Chatbot1 and Chatbot2, respectively. The “Avg.” in column name represents the overall contradiction rate of each chatbot. The “Avg.” in row name can be regarded as the ability to induce other chatbots to redeclare about the facts or opinions that are likely to be contradictory.
In the automatic evaluation, the ranking of consistency for the chatbots is Plato $\textgreater$ DialoGPT $\textgreater$ DialoFlow $\textgreater$ Blender, which is the same with expert evaluation. The Blender reaches the highest contradiction rate.

\noindent \textbf{Human Evaluation Results.}
We list the evaluation results in the bottom of Table \ref{tab:avg_contradiction_rate}. As we expected, BL obtains the highest contradiction rate. Meantime,  human evaluation also provides the same consistency ranking: Plato $\textgreater$ DialoGPT $\textgreater$ DialoFlow $\textgreater$ Blender as before. 
% The contradiction rates in human evaluation are significantly higher than that in the automatic evaluation, which reflects the limitation of the current neural model and it's important to adjust the threshhold $\tau$. We will make a deep discussion in the next section. 

\noindent \textbf{Summary.}
Both automatic evaluation and human evaluation in our framework can give the same performance ranking with the expert, which demonstrates that our framework is general and can effectively evaluate the consistency of chatbots.

\subsection{Time Efficiency}

Prior consistency evaluation methods with human-bot interaction are costly and take up a long time, which seriously slows down the development cycle of dialogue systems. In this section, we try to illustrate that our proposed \textit{Addressing Inquiries about History} framework is time and cost efficient and can help the evolution process of dialogue systems compared to the other methods. 

\begin{table}[t]
		\centering
		\begin{tabular}{c|c|c}
			\toprule[1pt]
			\bf{Method} & Time (Sec) & Contradiction \\
			\hline
			\bf{AIH (Auto)} & - + - & 1.56 \\
			\bf{AIH (Human)} & - + 24 & 1.69 \\
			\bf{Human-bot} & 246 + 59 & 0.50 \\
			\bottomrule[1pt]
		\end{tabular}
		\caption{The time efficiency of our proposed \textit{Addressing Inquiries about History} framework and traditional evaluation method with human-bot interaction. “Time” represents the time to create a conversation and the time to annotate the contradictions in a conversation. “Contradiction” denotes the average number of contradictions per conversation (average 15 turns). Dash line denotes the time can be ignored.}
		\label{tab:time-efficiency}
		%\ {-1\baselineskip}
\end{table}

As shown in Table  \ref{tab:time-efficiency}, we compare the time cost on two aspects: (i) the time to create inquires, and (ii) the time to detect contradictions in conversation. \textit{Addressing Inquiries about History} framework is based on the bot-bot conversation so that the time to create conversation can be ignored, while the human-bot conversation takes around 4 minutes per conversation. For the contradiction detection time, prior methods take around 1 minute considering the whole dialogue, while in our proposed framework, it is only about 24 seconds for human annotation or ignored for automatic evaluation.
Besides, we also compare the number of contradictions per conversation. As shown in Table \ref{tab:time-efficiency}, in our framework, the chatbots generate much more contradictions than those in prior methods. The detected contradictions are helpful for the chatbot developer to further improve the consistency of the chatbot. 

\noindent \textbf{Summary.} Our proposed framework can detect more contradictions with much less time than previous methods. Correspondingly, \textit{Addressing Inquiries} framework will accelerate the evolution process of consistency of chatbots.

\subsection{Ranking Stability}

\begin{figure}[t]
    \centering
    \includegraphics[width=0.48 \textwidth]{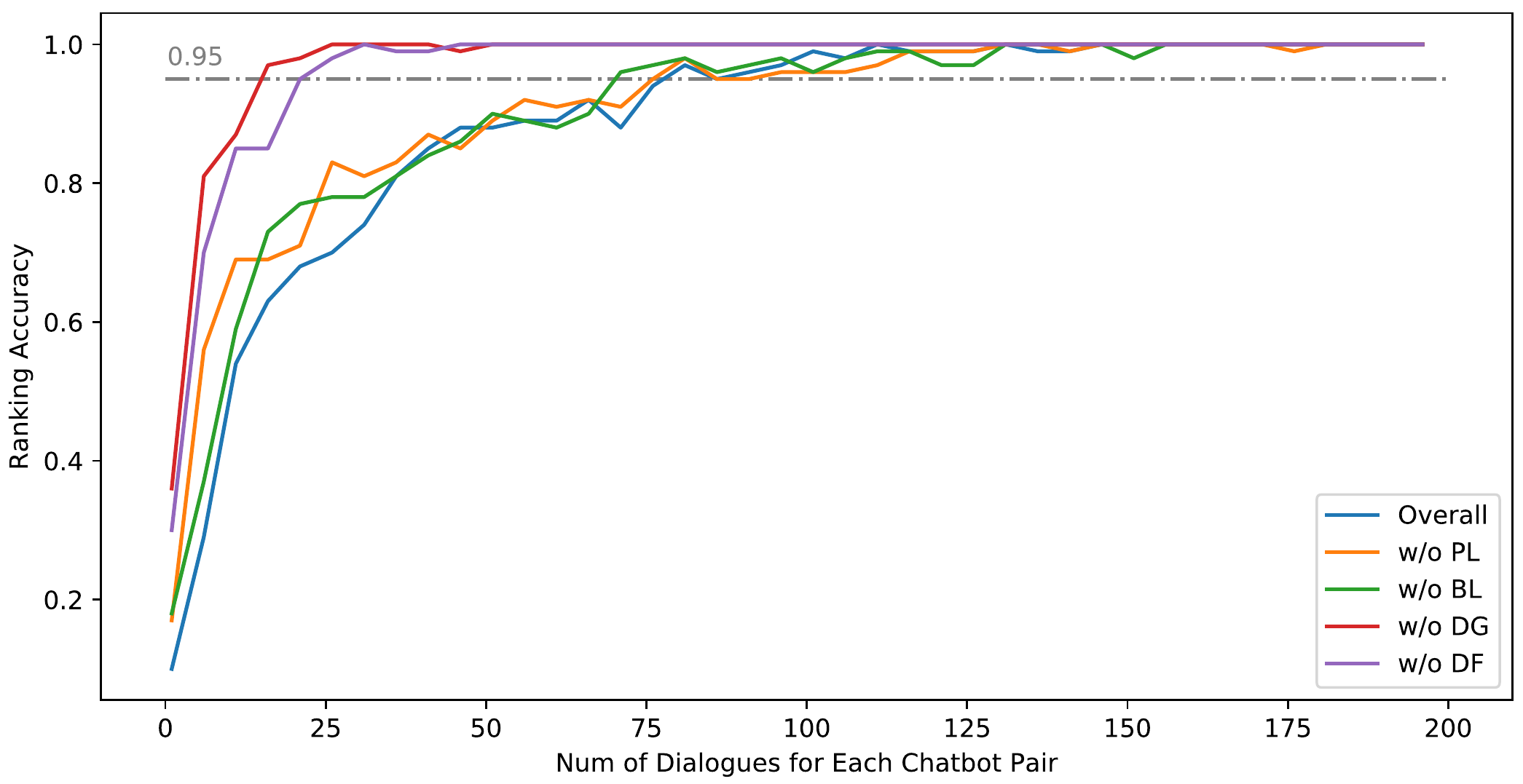}
    \caption{Ranking stability experiments. The x-axis denotes the number of conversations for each chatbot pair. The y-axis denotes the rate achieving the same ranking with the experts.}
    \label{fig:reliability}
\end{figure}

One key requirement for an evaluation framework is that repeated executions of the procedure result in the same outcomes. We measure how many conversations between each chatbot pair are required to guarantee a stable ranking. We randomly sample $\hat S$ conversations for each chatbot pair and compute the consistency ranking using automatic evaluation, where $\hat S \in \{1,\cdots,200\}$. We repeat this sub-sampling procedure 1000 times and compute the accuracy of achieving the same ranking with the expert ranking. 
As shown in Figure \ref{fig:reliability}, when $\hat S \textgreater 100$, the ranking results of the four chatbots are the same with the expert in 95\% cases and guarantee a stable ranking. We also do more in-depth analysis. The ranking stability depends on the significance of ranking. Table \ref{tab:expert_score} shows the consistency scores of DialoGPT and DialoFlow are close. We applied a leave-one-out stability analysis, in which we drop one chatbot. Figure \ref{fig:reliability} shows that when leaving one between DialoGPT or DialoFlow out, the stability is achieved with $\hat S=50$ dialogues. 

\noindent \textbf{Summary.}  The number of conversations needed for a stable evaluation in AIH framework is dependent on the chatbots to be tested, and more conversations usually lead to more stable evaluation. In general cases, 75 conversations are enough to get a valid contradiction detection. 

% \subsection{}

\section{Further Investigation}
In this section, we will further discuss the effectiveness of three parts in our framework containing question generation, contradiction detection, and human annotation evaluation. 

\subsection{Question Generation} 
Since a suitable question is necessary for the inquiry stage under our AIH framework, we make an in-depth analysis about the characters of question generation during inquiry stage.

\begin{table}[t]
		\centering
		\begin{tabular}{c|cccc|c}
			\toprule[1pt]
			\multicolumn{6}{c}{Number of Questions} \\
			\hline
			 & \bf BL & \bf PL & \bf DG & \bf DF & \bf Avg. \\
			\hline
			\bf{BL} & 6.54 & 6.13 & 2.62 & 5.12 & \bf 5.10 \\
			\bf{PL} & 6.54 & 5.34 & 1.98 & 4.36 & 4.55 \\
			\bf{DG} & 6.25 & 4.45 & 1.67 & 3.79 & 4.04 \\
			\bf{DF} & 6.15 & 5.79 & 2.25 & 4.21 & 4.60 \\
			\hline
			\bf{Avg.} & \bf 6.37 & 5.42 & 2.13 & 4.37 & 4.57 \\
			\toprule[1pt]
			\multicolumn{6}{c}{Number of Contradictions ($\tau=0.15$)} \\
			\hline
			 & \bf BL & \bf PL & \bf DG & \bf DF & \bf Avg. \\
			\hline
			\bf{BL} & 2.61 & 1.28 & 1.61 & 1.50 & \bf 1.74 \\
			\bf{PL} & 2.82 & 1.40 & 0.58 & 1.56 & 1.53 \\
			\bf{DG} & 2.66 & 1.12 & 0.57 & 1.31 & 1.38 \\
			\bf{DF} & 2.63 & 1.53 & 0.77 & 1.56 & 1.61 \\
			\hline
			\bf{Avg.} & \bf 2.73 & 1.38 & 0.69 & 1.57 & 1.56 \\
% 			\multicolumn{6}{c}{Number of Contradictions ($\tau=0.5$)} \\
% 			\hline
% 			 & \bf BL & \bf PL & \bf DO & \bf DF & \bf Avg. \\
% 			\hline
% 			\bf{BL} & 1.93 & 1.02 & 1.64 & 1.36 & \bf 1.51 \\
% 			\bf{PL} & 1.84 & 1.06 & 1.14 & 1.05 & 1.28 \\
% 			\bf{DO} & 1.39 & 0.57 & 0.54 & 0.67 & 0.8 \\
% 			\bf{DF} & 1.75 & 0.98 & 1.38 & 1.05 & 1.3 \\
% 			\hline
% 			\bf{Avg.} & \bf 1.73 & 0.87 & 1.16 & 1.03 & 1.19 \\
			\bottomrule[1pt]
		\end{tabular}
		\caption{Statistic of average number of inquiry pairs and the contradictions per conversation for each chatbot pair. The column name and the row name represent Chatbot1 and Chatbot2 respectively.}
		\label{tab:avg_amount}
		%\ {-1\baselineskip}
\end{table}

\noindent \textbf{Number of Questions and Contradictions} \\
We randomly sample 200 dialogues for each chatbot pair and make statistics on the average number of the inquiry pairs and contradictions per conversation. As shown in Table \ref{tab:avg_amount}, there are 4.57 inquiry pairs per conversation on average. 
There are 6.37 and 5.10 inquiry pairs per conversation when the Blender acts as Chatbot2 and serves as Chatbot1, respectively, which are both highest among all chatbots. 
The number of inquiry pairs reveals that the Blender can chat more about persona and facts, and the DialoGPT mentions these things less.
Table \ref{tab:avg_amount} also shows the number of contradictions per conversation. Similarly, the Blender makes the most contradictions and is the most likely to induce the chatbot interacting with it to redeclare facts or opinions that are likely to be contradictory.

\begin{table}[t]
		\centering
		\begin{tabular}{c|cccc|c}
			\toprule[1pt]
			\multicolumn{6}{c}{Question Appropriateness} \\
			\hline
			 & \bf BL & \bf PL & \bf DG & \bf DF & \bf Avg. \\
			\hline
			\bf{BL} & 0.932 & 0.960 & 0.922 & 0.936 & 0.938 \\
			\bf{PL} & 0.942 & 0.976 & 0.940 & 0.948 & 0.951 \\
			\bf{DG} & 0.784 & 0.870 & 0.928 & 0.882 & 0.866 \\
			\bf{DF} & 0.867 & 0.934 & 0.922 & 0.939 & 0.915 \\
			\hline
			\bf{Avg.} & 0.881 & 0.935 & 0.947 & 0.942 & 0.927 \\
			\bottomrule[1pt]
		\end{tabular}
		\caption{The appropriateness of the generated questions (Human evaluation).}
		\label{tab:question_appro}
		%\ {-1\baselineskip}
\end{table}

\noindent \textbf{Question Appropriateness} \\
We  analyze the appropriateness of the generated questions. We randomly sample 50 dialogues from each chatbot pair and ask human annotators to decide if the generated questions are appropriate based on the provided context (0/1). As shown in Table \ref{tab:question_appro}, the overall appropriateness score is about 0.93, which reveals that our question generation strategy is simple yet highly effective. We further study the wrong questions and find that most of them can be attributed to that the general question generation model can not work well in the dialogue context. We leave the better question generation task in open-domain dialogue for future work.

\subsection{Effect of Contradiction Threshold $\tau$}

We evaluate the effect of hyper-parameter $\tau$ in Equ.\ref{tau}, and the results are reported in Table \ref{tab:tau}. We compute the F1 score and Pearson correlation between the automatic evaluation results and the human annotations under different $\tau$. We can make the following observations: (i) When $\tau=0.15$, the Pearson correlation and F1 score reaches the highest. Thus we choose $\tau=0.15$ in our main experiments. (ii) The highest F1 score is 0.655, and the highest Pearson correlation is 0.436, which is a moderate correlation.
The observations reveal that there is a big gap between automatic evaluation and human evaluation, though the contradiction rate is similar. We consider that it is because the NLI model we employ is trained on the general domain rather than the dialogue domain, so there are lots of reference problems that can not deal with well. 

\begin{table}[t]
		\centering
		\begin{tabular}{c|ccc}
			\toprule[1pt]
			 & \bf CR & \bf F1 & $r$ \\
			\hline
			\bf{$\tau=0.1$} & \bf 0.401 & 0.650 & 0.430 \\
			\bf{$\tau=0.15$} & 0.364 & \bf 0.655 & \bf 0.436 \\
			\bf{$\tau=0.3$} & 0.287 & 0.606 & 0.423\\
			\bf{$\tau=0.5$} & 0.235 & 0.572 & 0.421 \\
			\bottomrule[1pt]
		\end{tabular}
		\caption{The analysis of threshold $\tau$. CR means contradiction rate. $r$ denotes the Pearson correlation. Pearson correlation and F1 score are measured with human annotations.  }
		\label{tab:tau}
\end{table}

\begin{table}[t]
		\centering
		\begin{tabular}{c|cccc|c}
			\toprule[1pt]
			\multicolumn{6}{c}{Inter-Annotator Agreement} \\
			\hline
			 & \bf BL & \bf PL & \bf DG & \bf DF & \bf Avg. \\
			\hline
			\bf{AIH} & 0.818 & 0.817 & 0.812 & 0.807 & 0.814 \\
			\bottomrule[1pt]
		\end{tabular}
		\caption{We analyse the inter-annotator agreement of the human evaluation in our proposed AIH framework. The correlation is measured by correlating each annotation with overall decision.}
		\label{tab:agreement}
\end{table}

\subsection{Inter-Annotator Agreement} 
To investigate the quality of human annotation, we compute the inter-annotator agreements, i.e., the correlation between each annotation and the overall decision is measured. The Pearson correlation for each chatbot is shown in Table \ref{tab:agreement}. The inter-annotator agreement is high for all chatbots, suggesting that the evaluation instructions are well-understood by the annotators.

\section{Conclusion and Future Work}
In this work, we introduced the \textit{Addressing Inquiries about History} (AIH), an effective and practical framework for open-domain chatbot consistency evaluation. AIH works by inserting questions about the mentioned  facts and opinions in the history into the bot-bot conversation and employing human annotators or neural models to evaluate whether the responses are consistent or not. Based on this, we can rank different chatbots accurately and 
efficiently. We show that our framework can effectively evaluate the consistency of chatbots and the evaluation results well correlate with experts. Also, our framework is cost and time-efficient and can not only give an overall consistency score but also provide exactly the contractions, which can accelerate the evolution process of chatbots. 

As in this work, we only focus on the contradictions about entities, and future work can improve the inquirer module and explore more kinds of contradictions. Besides, future work should also develop a more effective contradiction recognition module in the dialogue domain, while in this work we just exploit the general Natural Language Inference model to detect contradictions.
The non-consistency problem is serious in current open-domain chatbots.  We hope our work could facilitate and provide guidelines for future work on developing self-consistent open-domain chatbots.

\section*{Acknowledgement}
We sincerely thank the anonymous reviewers for their thorough reviewing and valuable suggestions. This work is supported by National Key R\&D Program of China (NO. 2018AAA0102502).

\bibliographystyle{acl_natbib}
\bibliography{anthology,acl2021}

\end{document}